\documentclass[conference]{IEEEtran}
\IEEEoverridecommandlockouts
\usepackage{cite}
\usepackage{amsmath,amssymb,amsfonts}
\usepackage{algorithmic}
\usepackage{graphicx}
\usepackage{textcomp}
\usepackage{xcolor}
\def\BibTeX{{\rm B\kern-.05em{\sc i\kern-.025em b}\kern-.08em
    T\kern-.1667em\lower.7ex\hbox{E}\kern-.125emX}}

\usepackage{booktabs}
\usepackage{multirow}
\usepackage{hyperref}
\hypersetup{hidelinks=true}
\usepackage{latexsym}
\usepackage{verbatim}

\usepackage{enumitem}

\begin{document}
\title{Large Language Models for Drug Overdose Prediction from Longitudinal Medical Records}
\author{Md Sultan Al Nahian\IEEEauthorrefmark{2}, Chris Delcher\IEEEauthorrefmark{1}, Daniel Harris\IEEEauthorrefmark{1}, Peter Akpunonu\IEEEauthorrefmark{1} and Ramakanth Kavuluru\IEEEauthorrefmark{1}\\
\IEEEauthorrefmark{1}University of Kentucky, Lexington, KY USA.
\IEEEauthorrefmark{2}Penn State Harrisburg, Middletown, PA USA.
}

\maketitle

\begin{abstract}
The ability to predict drug overdose risk from a patient's medical records is crucial for timely intervention and prevention. Traditional machine learning models have shown promise in analyzing longitudinal medical records for this task. However, recent advancements in large language models (LLMs) offer an opportunity to enhance prediction performance by leveraging their ability to process long textual data and their inherent prior knowledge across diverse tasks. In this study, we assess the effectiveness of OpenAI's GPT-4o and Google's MedGemma-27B LLMs in predicting drug overdose events using patients' longitudinal insurance claims records. We evaluate their performance in both fine-tuned and zero-shot settings, comparing them to strong traditional machine learning methods as baselines. Our results show that fine-tuned GPT-4o outperforms the traditional baselines, achieving F1 scores of 84.5 and 84.8 for the 7-day and 30-day prediction windows, compared to 78.9 and 77.3 for XGBoost, the strongest baseline. In addition, both LLMs can predict overdose risk in a zero-shot setting without any task-specific training. These findings highlight the potential of LLMs in clinical decision support for drug overdose risk prediction, but need prospective validation for real world deployment.

\end{abstract}


\section{Introduction}
Drug overdose (OD) is a major public health crisis in the United States, leading to a substantial number of emergency medical interventions and fatalities each year. According to the Centers for Disease Control and Prevention (CDC), drug overdoses claimed  107,941~\cite{Spencer2023Drug} lives in the U.S. in 2022, highlighting the urgent need for effective prevention and intervention strategies. 
Besides fatal outcomes and lost quality of life for patients, the misuse of prescription medications, illicit drugs, and polysubstance abuse has placed an immense burden on healthcare systems, emergency responders, and policymakers. 
Identifying individuals at risk early can facilitate timely interventions, such as targeted clinical assessments, behavioral support, and prescription monitoring, thereby reducing the likelihood of fatal outcomes.

Traditionally, drug overdose risk assessment has relied on clinical judgment, patient self-reporting, and structured risk assessment tools based on demographic, behavioral, and medical history data. One widely used tool is the Prescription Drug Monitoring Program (PDMP)~\cite{finley2017evaluating}, an electronic database that tracks prescriptions for controlled substances within a state. Authorized users, including physicians, pharmacists, law enforcement, and public health officials, can query PDMPs to identify patterns of potentially harmful or unlawful prescription drug use. However, PDMPs often suffer from incomplete data, and their limited interstate data sharing makes it difficult to monitor patients who obtain medications across state lines. Moreover, delays in data entry can further contribute to missing information, impeding safe and informed decision-making~\cite{marie2023barriers}. While PDMPs and toxicology screenings provide valuable insights into potential overdose risks, these methods are primarily reactive rather than proactive. They also rely on clinicians manually reviewing a patient’s medical history, which may lead to overlooked patterns and inconsistencies. As a result, traditional approaches may fail to capture the complex, longitudinal factors that contribute to overdose risk, highlighting the need for a data-driven automated process.

To overcome the limitations of manual process, in recent years, machine learning (ML) approaches have emerged as promising tools for improving overdose prediction by leveraging large-scale electronic health records (EHRs), insurance claims data, and prescription histories. 
Traditional ML models, such as logistic regression, random forests, and deep learning methods~\cite{lo2019evaluation, che2018deep, ma2017dipole}, have demonstrated the ability to identify high-risk individuals by analyzing factors such as prior diagnoses, medication prescriptions, healthcare utilization patterns, and social determinants of health. These models can process structured and unstructured medical data, uncovering hidden correlations that may be overlooked in traditional assessments. However, many existing ML approaches require extensive feature engineering and very often struggle to capture the temporal dependencies in sequential patient visits.

The recent advancements of large language models (LLMs)~\cite{llm2020} present a promising opportunity to overcome the limitations of traditional machine learning models in predicting drug overdose risk. LLMs are pretrained on vast amounts of data, including biomedical and clinical texts, enabling them to \textit{understand} complex medical information which may give them the ability to generate informed decisions based on a patient's medical history~\cite{wei2022emergent}. Unlike traditional ML models that require extensive feature engineering, LLMs may leverage their contextual understanding to process sequential medical events more effectively~\cite{liu2023large, hegselmann2023tabllm}. Additionally, these models possess prior knowledge from their pretraining, allowing them to make informed predictions across various tasks without the need for task-specific fine-tuning. In this study, we explore the potential of LLMs in predicting drug overdose events from the longitudinal medical history of the patient as captured in insurance claims.

Specifically, we investigate the capability of OpenAI's GPT-4o and Google's MedGemma-27B-text-it~\cite{sellergren2025medgemma} in predicting drug overdose events using the Merative MarketScan (formerly known as Truven) dataset, which includes patient diagnoses, procedure records, medication history, and demographics. We framed overdose prediction as a sequence modeling problem, where the model processed a patient's past medical visits to estimate the likelihood of a drug overdose event within two predefined time windows: next 7 days and 30 days. To achieve this, we designed different prompting templates to provide longitudinal visit information along with task instructions to the models. To assess the effectiveness of LLM-based approaches, we conducted both zero-shot inference, where the model was not trained with any task-specific data, and fine-tuning experiments, where the model was fine-tuned with such data. We then compared their performance against traditional machine learning models, including Random Forest and XGBoost, two widely used methods for structured data~\cite{ellis2019predicting, ramirez2025systematic}. 
Our analysis provides insights into the advantages and limitations of LLMs in clinical prediction tasks, particularly in modeling complex longitudinal patient data. The code used in this project is available at: \url{https://github.com/bionlproc/overdose_pred}.


\label{sec:introduction}


\section{Datasets}
For this study, we used data from the Merative MarketScan Research Databases \cite{marketscan2025} (formerly known as Truven). This comprehensive database contains de-identified, patient-specific health information derived from insurance claims, covering a large and diverse population. The dataset includes key healthcare data elements such as patient demographics, diagnoses, procedures, prescribed medications, and encounters. It provides a longitudinal view of patient encounters within the healthcare system, that enables in-depth analysis of healthcare trends, treatment patterns, and patient outcomes.

\subsection{Cohort Selection}
\label{sec-cohort-select}
In this study, we used the Merative MarketScan data spanning three years (2020-2022). The study cohorts include patients who: (1) are at least 18 years old, and (2) have at least 12 months of continuous data, with a minimum of five medical events (e.g., diagnoses, procedures, or prescriptions).

We extracted the cohorts based on these criteria and categorized them into two groups: case and control.
\begin{itemize}
    \item Case cohort: This cohort includes patients diagnosed with any drug overdose or poisoning event, identified using ICD-9 and ICD-10 codes.
    In ICD-10-CM, drug poisoning cases are classified under codes T36–T50, which cover poisoning, adverse effects, and underdosing of drugs, medicaments, and biological substances. These codes are frequently used in research to identify drug overdose cases~\cite{di2018drug, snow2021descriptive}.
    To ensure that the case cohort focuses exclusively on overdose cases, we first extracted all records with ICD-10-CM codes T36–T50 and then excluded cases related to adverse effects and underdosing. Specifically, cases with a 5th character of ‘5’ or ‘6’ were removed (e.g., T40.2X5, T40.2X6), as these indicate adverse effects and underdosing, respectively.

    For ICD-9-based identification, we included cases with the following codes: 965, 968, 969, 970, E850, E853, E854, and E858.

    
    \item Control cohort: The control cohort consists of patients who are not part of the case cohorts and have no recorded history of overdose diagnosis. This group also includes patients with drug poisoning cases due to underdosing or adverse effects but explicitly excludes those with documented overdose incidents.

    Additionally, the control cohort includes a subset called the ``exposed cohort." This group consists of patients who have been exposed to opioids or stimulants but have not experienced an overdose. Since the use of opioids or stimulants usually increases the risk of overdose~\cite{palis2022concurrent, bohnert2019understanding}, including these patients presents a unique challenge for the model in predicting overdose events. This group helps improve the model’s ability to differentiate between individuals who are exposed to opioids or stimulants but do not overdose, and those who do. By including exposed patients in the control cohort, we aim to make the model more robust, ensuring it can accurately assess overdose risk in high-risk populations.

    To identify exposed cohorts, we used prescription records and diagnosis data. Patients were included in this cohort if they met either of the following criteria:
    \begin{itemize}
        \item They had at least one prescription for opioids or stimulants, or
        \item They had a diagnosis of opioid or stimulant use disorder but no recorded overdose event.
    \end{itemize}

    For prescription-based identification, we used MEDISPAN root classes:
    \begin{itemize}
        \item ``ADHD/Anti-Narcolepsy/Anti-Obesity/Anorexiant Agents" (for stimulant prescriptions)
        \item ``Analgesics – Opioid" (for opioid prescriptions)
 
    \end{itemize}

For diagnosis-based identification, we included patients with a diagnosis of opioid or stimulant use disorder based on ICD-9 or ICD-10 codes:
\begin{itemize}
    \item ICD-9 codes: 304.0x, 304.2X, 304.4X, 304.7X, 305.5X, 305.6X, 305.7X, 305.8X
    \item ICD-10 codes: F11.XXX, F14.XXX, F15.XXX 
\end{itemize}

\end{itemize}

\subsection{Data Pre-processing}
After extracting the cohorts from the tables, we preprocessed the data to prepare it for our predictive modeling. Since the goal is to predict the risk of future overdose events for the patients based on their medical history available up to the date when the prediction is made, the dataset needs to include only the visits that occurred on or before the cutoff date. Any data recorded after this cutoff date must be excluded to ensure the model reflects the real-world scenario of making predictions using only past and present information.

\begin{figure*}[t]
\centering{
   
  \includegraphics[width=0.85\linewidth]{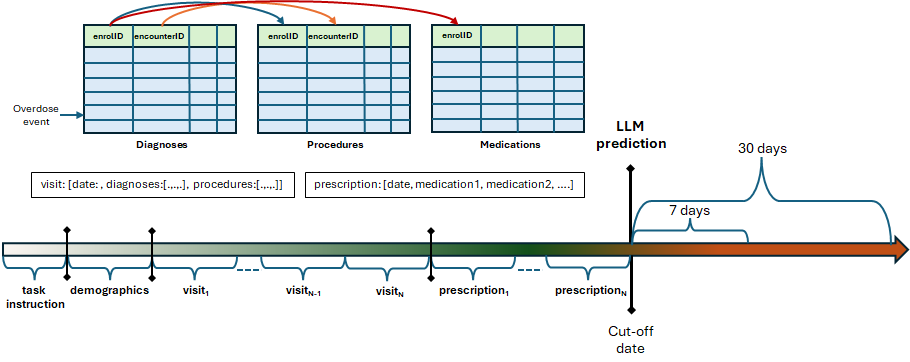} 
  
  \caption {The prompt consists of the task instruction for LLM followed by the patient's medical history. The medical history includes demographic information, visit/encounter details, and prescriptions. Both encounters and prescriptions are presented in chronological order, with the earliest first and the latest last. Each encounter contains diagnoses and procedures performed during the visit. The objective of the LLM is to predict the likelihood of a drug overdose within the next 7 and 30 days based on past encounters.}
        \label{fig:method}}

\end{figure*}

To accomplish this for the case cohort, we first identify the first recorded encounter of a drug overdose for each patient. Starting from this encounter, we trace back to the immediately preceding encounter and check if it falls within the specified prediction window (Figure~\ref{fig:method}). If the prior encounter's date lies within the prediction window, we designate it as the cutoff date for the patient’s medical history. All data following this cutoff date are excluded, ensuring that the dataset reflects the patient’s condition before the potential overdose event.
For example, with a 7-day prediction window, if a patient’s first overdose encounter occurs on July 30, 2022, and the immediately preceding encounter is on July 25, 2022, this prior date qualifies as the cutoff date since it falls within the 7-day window. The patient’s medical history up to July 25, 2022, is included as a positive case in the cohort, and the prediction task is to assess the risk of an overdose within the next 7 days. 

For the control and exposed cohorts, we focus on the most recent visit and examine the encounter immediately preceding it. If this preceding encounter falls within the prediction window, its date is designated as the cutoff date. All visit information up to this cutoff date is included in the dataset as a negative example. This ensures that the data represents patients who, at that point, have not experienced an overdose. Using the encounter just before the most recent visit as the cutoff ensures that the negative or control cases align with the temporal structure of the positive cases.


\section{Methods}
In this study, we aim to predict the risk of drug overdose from the longitudinal healthcare data of patients by using LLMs. We compare our results with baseline methods to show the effectiveness of using LLMs. In this section, we discuss the baseline methods and the proposed LLM-based method. 
\subsection{Baseline Methods}

For baselines, we used traditional machine learning algorithms, random forest and XGBoost, which are commonly used for structured data. Since these algorithms require static  input features, we derived a fixed set of features from the dataset. We included primary and secondary diagnosis codes, procedure codes, drug names, and drug therapeutic names along with their strength and routes of administration as the feature groups.

To define the feature set, we identified the most frequent values for each feature type, selecting items that appeared in at least 50 visits in the training dataset. This process resulted in a total of 3,700 features. For each patient, we constructed a feature vector by calculating the occurrences of these selected feature items across all visits. These feature vectors were then used to train the baseline models.

\subsection{Large Language Models}
\label{sec-meth}

Since LLMs are designed to process text-based inputs, and the MarketScan database is structured in a tabular format, we need to transform the longitudinal structured medical data into a format that LLMs can interpret effectively.
Several studies have demonstrated that converting structured tabular data into text enables LLMs to process it more effectively. Various serialization techniques have been proposed for this transformation. A common approach is to represent the tabular data in a programming language readable data structure, such as a Pandas DataFrame~\cite{singha_dfloader_2023tabular}, HTML code~\cite{singha_dfloader_2023tabular, sui2023tap4llm}, or a data matrix represented as a list of lists~\cite{singha_dfloader_2023tabular}. Other widely used methods include converting the data into JSON format~\cite{sui2024table}, key-value pairs~\cite{wang2023meditab}, or delimiter-separated values (e.g., CSV or TSV)~\cite{narayan2022can}. Additionally, some studies have explored converting tabular data into natural language sentences using predefined templates~\cite{hegselmann2023tabllm, gong2020tablegpt}.
In our study, we chose JSON format for data conversion due to its effectiveness, simplicity, and structured representation.

From the Merative MarketScan Database, we extracted four primary tables: diagnosis, procedure, encounter, and prescription, where each row represents a single record of an encounter (visit). For example, a row in the diagnosis table contains the detail of one diagnosis from an encounter, while the procedure table includes one procedure, and so on. Since a single encounter may involve multiple diagnoses, procedures, and other details, the information for an encounter spans multiple rows across these tables. To aggregate all details of a particular encounter, we used the Encounter ID, a unique identifier in the diagnosis, procedure, and encounter tables. Using this identifier, all rows in the diagnosis table associated with a specific encounter were linked to consolidate the diagnosis details. Similarly, rows in the procedure and encounter tables were linked and aggregated using the same Encounter ID, providing a combined record of all diagnoses, procedures, and encounter information for a single visit.

Each patient is uniquely identified by the Enrol ID, which is consistent across all tables. Using this identifier, we retrieved all encounters associated with a patient and aggregated the details of each visit using the process described above. The resulting visit-specific records for the patient were then arranged in chronological order, with earlier visits appearing first and the most recent visits at the end. 
This temporal arrangement preserved the progression of a patient's medical history over time, enabling a representation of their longitudinal care data.



In contrast to other tables, prescriptions are not linked to any specific encounter; rather, they are linked to individual patients and dates.
Therefore, to integrate prescription data, we extracted all prescriptions for each patient and arranged them chronologically using the ``fill date''. Each prescription was represented as a key-value pair, similar to the diagnosis and procedure details. 
The sequentially ordered prescription data was then appended to the pre-processed diagnosis, procedure, and encounter information for each patient. This ensured a unified and comprehensive representation of both visit-specific and medication information.
Finally, all preprocessed visits and prescriptions data for each patient were concatenated into a single text sequence, providing a detailed and temporally structured view of their healthcare history. This format allows LLMs to effectively leverage their natural language understanding capabilities to analyze the data and predict the risk of drug overdose.

To represent patient data as text sequences for LLMs, we used two formats: (1) detailed visit information and (2) statistical summaries of visits. Each of these formats was further structured in two ways: (a) using detailed descriptions and (b) using original medical codes.
\begin{enumerate}
    \item \textbf{Detailed visit information:} In this format, we converted each visit's details, prescriptions, and patient demographics into text. 
    \begin{itemize}
    
    \item \textit{Detailed descriptions:} In this structure, we replaced medical codes and database field names with their corresponding natural language descriptions to leverage LLMs' grasp on textual data. We used the corresponding description of the specific ICD-10 codes representing the diagnosis,   the descriptions of CPT/HCPCS and ICD-10-PCS procedure codes used for procedures, and so on. For example, instead of using the ICD-10 diagnosis code N18.9, which represents Chronic kidney disease, unspecified (CKD), we included its description: ``Chronic kidney disease, unspecified (CKD).'' Similarly, for database fields, we replaced technical names such as ``DIAG\_CD'' with human-readable labels ``diagnosis code''. We hypothesized that providing detailed natural language descriptions would help LLMs better understand the clinical context of a patient's medical history.
    \item \textit{Original medical codes:} In this format, we retained the original medical codes without converting them into natural language descriptions. For instance, the diagnosis code N18.9 was used as-is, rather than its description.
The use of original medical codes serves two purposes:
\textit{1.	Compact representation:} It reduces the length of the text sequence, which can be advantageous for processing efficiency.
\textit{2.	Capability testing:} It assesses whether LLMs can interpret and make predictions directly from standardized medical codes without additional context. 
\end{itemize}
    \item \textbf{Statistical summaries of visits:}     Unlike the first format, which represented each visit as text in a key-value format, this approach summarizes patient data statistically. It aggregates data from all visits of a patient into a single feature vector. This is the same feature vector used in our baseline methods, where we counted the occurrences of each feature (e.g., diagnosis codes, procedure codes) across all visits for a patient.
To make this format compatible with LLMs, we converted the feature vector into text by transforming field names and numerical values into natural language. This created a concise, high-level summary of a patient’s medical history. Like in the \textit{detailed visit information} format, we represented field names in two ways: using descriptive labels and their original medical codes.
\end{enumerate}
These different input formats result in four different representations, which we named as, 1) Detailed visit - descriptive, 2) Detailed visit - medical code, 3) Summarized visit - descriptive and 4) Summarized visit - medical code.

\section{Experimental Setup}
\label{Sec:experimental_setup}
In this study, we investigated the potential of LLMs to predict drug overdose events using longitudinal medical records of patients. For this purpose, we used GPT-4o (gpt-4o-2024-08-06) and MedGemma-27B-text-it as the LLMs. GPT-4o is a general-purpose large language model with strong performance across diverse tasks, while MedGemma is a medically specialized open-weight model developed by Google, pretrained on large-scale biomedical and clinical text corpora. Including MedGemma-27B-text-it in our evaluation allows us to assess whether domain-specific pretraining on medical text confers an advantage in clinical prediction tasks such as overdose risk assessment, and to compare an open-weight model against a proprietary one. To assess the performance of these models, we compared them against traditional machine learning models, which served as baselines. Specifically, we selected Random Forest and XGBoost as the baseline models due to their proven effectiveness and strong performance in classification tasks involving structured tabular data.

For GPT-4o, we tested three temperature values (0.5, 0.8, and 1.0) and found that 0.5 provided the best performance; this setting was therefore used for all GPT-4o experiments. For MedGemma-27B-text-it, we used the model's default inference settings for zero-shot evaluation. For fine-tuning the MedGemma model, we applied QLoRA (Quantized Low-Rank Adaptation)~\cite{dettmers2023qlora} to reduce memory requirements while attempting to preserve predictive performance. 
We fine-tuned all linear projection layers, as these layers contribute most directly to the model's prediction performance.
We experimented with LoRA rank values of r = 8, 16, and 32, and found that r = 32 yielded the best results. 
We therefore adopted r = 32 with a scaling factor of $\alpha$ = 64 for all fine-tuning experiments, resulting in approximately 227 million trainable parameters, which represents 0.83\% of the total parameters of the MedGemma-27B model. 
Both models were evaluated for two prediction windows: 7 days and 30 days. In the 7-day window, the model predicts whether a patient is at risk of overdose within the next 7 days, while in the 30-day window, it predicts the risk within the next 30 days. 


For each prediction window, we created separate datasets, each containing 900 samples for training, 900 for validation, and 900 for testing. Each set consists of 300 case cohort samples (overdose) and 600 control cohort samples (non-overdose), with 300 instances (50\%) of the control cohort being from the exposed group. All results reported in this paper are based on the test sets.
Note that for MedGemma, fine-tuning experiments were conducted only for the summarized visit prompt formats. This is because the detailed visit formats produce very long input sequences, which exceeded the GPU memory capacity when fine-tuning the 27B-parameter model even with QLoRA. For the zero-shot setting, where no gradient computation is required, MedGemma was evaluated across all four prompt formats.

\begin{table}[h]
    \centering
    \begin{tabular}{cccccc}
        \toprule
        Pred.~window & Model & P & R & Spec & F1 \\
        \midrule
        \multirow{2}{*}{7 days} & Random Forest & 88.89 & 66.67 & 95.83& 76.19 \\
        {} & XGBoost & 85.88 & 73.00 & 94.00 & 78.92 \\
        \midrule
        \multirow{2}{*}{30 days} & Random Forest & 84.02 & 61.33 &93.67 & 70.91 \\
        {} & XGBoost & 85.77 & 70.33 &95.50 & 77.29 \\

        \bottomrule
    \end{tabular}
\caption{Baseline performances for overdose prediction}
\label{table:result_baseline}
\end{table}

\begin{table*}[h]
    \centering
    \begin{tabular}{cccccccccc}
        \toprule
        \multirow{3}{*}{Pred.~window} & \multirow{3}{*}{Prompt Type} & \multicolumn{8}{c}{Model}\\
        \cmidrule{3-10}
        & & \multicolumn{4}{c}{GPT-4o} & \multicolumn{4}{c}{MedGemma-27B} \\
        \cmidrule{3-10}
        & & Precision & Recall & Specificity & F1-score & Precision & Recall & Specificity & F1-score\\
        
        \midrule
        \multirow{4}{*}{7 days} & Detailed visit - descriptive & 57.68 & 51.33 & 81.17 & \textbf{54.32} & 37.10 & 82.00 & 30.50 & 51.09\\

        {} & Detailed visit - med. code & 54.87 & 50.67 & 79.17 & \underline{52.69} & 36.19 &70.33 & 38.00 &47.79\\
        {} & Summarized visit - descriptive & 57.08 & 45.67 & 82.83 & 50.74 & 43.41 & 52.67 & 65.67& 47.59\\
        {} & Summarized visit - med. code & 53.66 & 44.00 & 81.00 & 48.35 & 41.88 & 49.00 & 66.00&45.16\\
        \midrule
        \multirow{4}{*}{30 days} & Detailed visit - descriptive & 58.54 & 56.00 & 80.17 & \textbf{57.24} &37.77&88.00&27.50&52.85\\
        {} & Detailed visit - med. code & 53.02 & 55.67 & 75.33 & \underline{54.31} &36.48&82.33&28.33&50.56\\
        {} & Summarized visit - descriptive & 55.75 & 42.00 & 83.33 & 47.91 &43.30&60.33&60.50&50.42\\
        {} & Summarized visit - med. code & 51.44 & 41.67 & 80.33 & 46.04 &37.50&52.00&56.67&43.58\\

        \bottomrule
    \end{tabular}
\caption{Results of Zero-shot overdose prediction with LLMs}
\label{table:LLM-zero-shot}
\end{table*}

\begin{table*}[h]
    \centering
    \begin{tabular}{cccccccccc}
        \toprule
        \multirow{3}{*}{Pred.~window} & \multirow{3}{*}{Prompt Type} & \multicolumn{8}{c}{Model}\\
        \cmidrule{3-10}
        & & \multicolumn{4}{c}{GPT-4o} & \multicolumn{4}{c}{MedGemma-27B} \\
        \cmidrule{3-10}
        & & Precision & Recall & Specificity & F1-score & Precision & Recall & Specificity & F1-score\\
        \midrule
        \multirow{4}{*}{7 days} & Detailed visit - descriptive & 74.1 & 65.67 & 89.00 & 69.99 &-&-&-&-\\
        {} & Detailed visit - med. code & 71.19 & 70.00 & 85.83 & 70.59 &-&-&-&-\\
        {} & Summarized visit - descriptive & 87.23 & 82.00 & 94.00 & \textbf{84.53} &89.62&54.67&96.83&67.91\\
        {} & Summarized visit - med. code & 89.47 & 79.33 & 95.33 & \underline{84.10} &86.70&58.67&95.50&69.98\\
        \midrule
        \multirow{4}{*}{30 days} & Detailed visit - descriptive & 74.05 & 71.33 & 87.50 & 72.62 &-&-&-&-\\

        {} & Detailed visit - med. code & 74.91 & 68.67 & 88.50 & 71.65 &-&-&-&-\\
        {} & Summarized visit - descriptive & 95.80 & 76.00 & 98.33 & \textbf{84.76} &87.75&59.67&95.83&71.04\\
        {} & Summarized visit - med. code & 88.24 & 80.00 & 94.67 & \underline{83.92} &100.0 &40.00 & 100.0&57.14\\

        \bottomrule
    \end{tabular}
\caption{Results of fine-tuned overdose prediction with LLMs}
\label{table:LLM-fine-tuned}
\end{table*}

\begin{table*}[h]
    \centering
    \begin{tabular}{cccccc}
        \toprule
        \multirow{3}{*}{Pred.~window} & \multirow{3}{*}{Prompt Type} & \multicolumn{4}{c}{Model}\\
        \cmidrule{3-6}
        & & \multicolumn{2}{c}{GPT-4o} & \multicolumn{2}{c}{MedGemma-27B} \\
        \cmidrule{3-6}
        & & Zero-shot & Fine-tune & Zero-shot & Fine-tune\\
        \midrule
        \multirow{4}{*}{7 days} & Detailed visit - descriptive & 0.38 &0.62 & - & -\\
        {} & Detailed visit - med. code & 0.35 & 0.61 & - & - \\
        {} & Summarized visit - descriptive & 0.32 & 0.77 & 0.30 & 0.56\\
        {} & Summarized visit - med. code & 0.28 & 0.77 & 0.25 & 0.61\\
        \midrule
        \multirow{4}{*}{30 days} & Detailed visit - descriptive & 0.41 & 0.64 & - & - \\
        {} & Detailed visit - med. code & 0.35 & 0.63 & - & - \\
        {} & Summarized visit - descriptive & 0.28 & 0.80 & 0.32 & 0.66\\
        {} & Summarized visit - med. code & 0.25 & 0.77 & 0.22 & 0.55\\

        \bottomrule
    \end{tabular}
\caption{Matthews Correlation Coefficient score of the models}
\label{table:mcc}
\end{table*}

 \section{Results}
 Table \ref{table:result_baseline} shows the performance of the baseline models, random forest and XGBoost, in predicting overdose events. Both models achieve F1 scores between 70 and 80, indicating strong baseline performance. Across both prediction windows, XGBoost consistently achieves higher F1 scores than random forest. While XGBoost identifies more positive overdose events, random forest offers greater precision, producing fewer false positives.

The results of LLM experiments are presented in Tables~\ref{table:LLM-zero-shot} and~\ref{table:LLM-fine-tuned}. Table~\ref{table:LLM-zero-shot} highlights the effectiveness of LLMs in predicting overdose events in the zero-shot setting. We conducted experiments using two prediction windows (7 days and 30 days) and evaluated performance across four different prompt templates. The results indicate that both models perform best when provided with detailed English descriptions of diagnosis and procedure codes and prescriptions, with this prompt format achieving both highest recall and precision, outperforming other prompt formats across both prediction windows.

In the second prompt format, we used original medical codes for diagnoses and procedures instead of their natural language descriptions. For GPT-4o, this format achieves F1 scores of 52.69 and 54.31 for the 7-day and 30-day prediction windows, respectively. Although performance is slightly lower than with natural language descriptions, this approach has a crucial advantage: by relying on original codes, the input prompt size is considerably reduced, which in turn lowers the computational cost of running the GPT-4o API. Table~\ref{table:LLM-zero-shot} also presents results for prompt formats that utilize summary statistics from patients' past visits, which we refer to as summarized visit formats. In this approach, aggregated statistics of diagnoses, procedures, and medications over the last n visits are provided in two variations: one using original medical codes and the other using their English descriptions. The summarized visit formats yield the lowest zero-shot F1 scores for GPT-4o (50.74 and 48.35 for the descriptive and medical code variants in the 7-day window), underscoring the importance of detailed, context-rich input for zero-shot prediction. The specificity of GPT-4o across all prompt formats ranges from 75 to 83, indicating its strong capability to correctly identify negative cases in the zero-shot setting.

For MedGemma-27B-text-it in the zero-shot setting, a markedly different pattern emerges. MedGemma achieves much higher recall than GPT-4o across all prompt formats, but at the cost of much lower precision and specificity. For example, with the detailed descriptive prompt in the 30-day window, MedGemma achieves a recall of 88.00 compared to GPT-4o's 56.00, but its specificity drops to just 27.50 versus GPT-4o's 80.17. This high-recall, low-precision and low-specificity pattern is consistent across all prompt types and both prediction windows, indicating that in the zero-shot setting MedGemma tends to over-predict overdose risk. This likely reflects its medical domain pretraining, which makes it highly sensitive to clinical risk signals but poorly calibrated for this specific binary classification task without task-specific fine-tuning. Among MedGemma's zero-shot results, the summarized visit formats yield notably higher specificity (65.67 and 66.00 for the 7-day window) compared to the detailed visit formats (30.50 and 38.00), suggesting that the aggregated representation partially mitigates MedGemma's tendency to over-predict.

Table~\ref{table:LLM-fine-tuned} presents the results of the fine-tuned models. For GPT-4o, fine-tuning significantly enhances performance across all prompt formats. In the 7-day prediction window, the F1 scores for the Detailed Visit - descriptive and Detailed Visit - med. Code formats are 69.99 and 70.59, respectively, representing improvements of 16 and 18 points over their zero-shot counterparts. Fine-tuning improves the model's ability to predict positive overdose events accurately while also enhancing its capability to correctly identify non-overdose cases, as reflected in the increases in both recall and specificity. The most significant improvement is observed when the model is fine-tuned using aggregated summary statistics of visit data for each patient. It is important to note that the baseline models were also trained on the same aggregated data for comparison. As shown in Table~\ref{table:LLM-fine-tuned}, the fine-tuned GPT-4o achieves an F1 score of 84.53 (7-day, summarized descriptive), representing an approximately 6-point gain over XGBoost (F1: 78.92), the best-performing baseline. Recall improves to 82.00, which is a 9-point gain over XGBoost, while specificity reaches 94.00, confirming that fine-tuned GPT-4o delivers a strong overall balance across all metrics.

For MedGemma-27B-text-it, fine-tuning was performed only on the summarized visit formats due to GPU memory constraints with the detailed visit formats. Fine-tuning largely corrects the over-prediction tendency observed in the zero-shot setting, producing large gains in both precision and specificity. In the 7-day window, precision increases to 89.62 and 86.70 for the summarized descriptive and medical code formats, respectively, compared to 43.41 and 41.88 in the zero-shot setting, and specificity rises to 96.83 and 95.50 from 65.67 and 66.00. Recall also improves in this window, though by a smaller margin, from 52.67 to 54.67 for the descriptive format and from 49.00 to 58.67 for the medical code format. In the 30-day window, the descriptive format follows a similar pattern, reaching a precision of 87.75 and a specificity of 95.83 while recall remains nearly unchanged (60.33 vs. 59.67). The medical code format behaves differently: it reaches perfect precision and specificity (100.0), but its recall drops from 52.00 to 40.00, indicating that the fine-tuned model predicts overdose only when it is highly confident and misses many positive cases. Overall, fine-tuning improves the F1 score in all four settings and transforms MedGemma from an over-predicting zero-shot model into a high-precision, high-specificity model, while its effect on recall is mixed (gains in the 7-day window, flat or lower recall in the 30-day window). Even so, GPT-4o achieves a more balanced trade-off across all metrics and remains the stronger fine-tuned model in both prediction windows.

To evaluate the classification quality of the models, we also report the Matthews Correlation Coefficient (MCC) in Table~\ref{table:mcc}. MCC ranges from -1 to +1, where 0 means the model is only guessing, and it considers both correct and incorrect predictions for both classes. The MCC results match what we see in the other metrics. Without any training, GPT-4o scores between 0.25 and 0.41, showing it can find real signal in the data on its own, while MedGemma scores a bit lower (0.22 to 0.32). After fine-tuning, both models improve substantially: GPT-4o reaches 0.77 and 0.80 on the summarized-descriptive format for the 7- and 30-day windows, and MedGemma rises to between 0.55 and 0.66. Overall, fine-tuned GPT-4o achieves the strongest performance, and the consistently positive MCC scores demonstrate that both models capture a meaningful relationship between a patient's claims history and their subsequent overdose risk.

 \section{Concluding Remarks}

 The ability to predict the risk of drug overdose is crucial in healthcare, as it enables timely intervention and has the potential to save lives. In this study, we evaluated the effectiveness of two large language models — GPT-4o and MedGemma-27B-text-it — in predicting drug overdose risk based on patients' longitudinal medical records, examining predictive performance for two time intervals: within 7 days and 30 days from the observation date. Our experimental results show that fine-tuned GPT-4o outperforms traditional baseline methods when provided with consolidated statistical summaries of prior diagnoses, procedures, and medications. In the zero-shot setting, MedGemma-27B-text-it achieves high recall across all prompt formats, making it a strong candidate for broad initial screening, though its low specificity means it must be used with awareness of its high false positive rate. Fine-tuning (feasible only for the summarized formats) substantially improves MedGemma’s precision and specificity, but its recall changes little and remains well below the 82–88\% achieved zero-shot with detailed visit formats, resulting in overall F1 scores below those of fine-tuned GPT-4o.. 
 These complementary performance profiles suggest that model selection should be guided by the clinical use case: zero-shot MedGemma for maximum sensitivity, and fine-tuned GPT-4o for a balanced precision-recall trade-off.
 The code used for this project is publicly available: \url{https://github.com/bionlproc/overdose_pred}.

\subsection{Additional Analyses in the Appendix}
 There are several important comparisons that were not discussed in the main portion of this paper. For instance, the influence of the length of the history (in terms of number of prior visits) available in the observation window was not discussed. On the surface it appears the longer the history, the better. But we investigate this with additional experiments in the Appendix.  
 As discussed in Section~\ref{sec-meth}, we used both descriptions of the medical codes and the raw codes themselves in different configurations of the experiments. In that section, we also talk about furnishing full longitudinal records vs aggregated summary statistics to the models. Although we briefly discussed this earlier, we go into additional details in the Appendix. 
 In Section~\ref{sec-cohort-select}, we introduced a so-called ``exposed'' cohort, a special subset of the negative cases with patients who were exposed to opioids/stimulants but did not have a coded overdose. We dig into model performances of this subset in the Appendix. 
 Finally, we also discuss cost comparisons of LLM API calls of several configurations at the end of the Appendix. 

 \subsection{Limitations}

Our approach also has certain limitations. 
First, we oversampled the overdose cases in our cohort to be around $33\%$, while in reality the prevalence of overdose in general population is $< 1\%$. Though oversampling is often necessary  in the training dataset to have enough learning signal, in our case, it also persists in the test dataset to ensure we have enough true overdose cases to test our methods. Our experiments shed light on the comparisons involving different configurations (codes vs textual descriptions, full history vs aggregate measures, length of history, zero-shot vs fine-tuning) involving two mainstream LLMs. That said, our results are not meant to offer decisive guidance on whether LLMs are ready for deployment at scale for overdose prediction in the real world. 
Second, in our current model, we utilized demographics, diagnoses, procedures, and medication records but did not include laboratory data, as it was not available in the Merative MarketScan (formerly known as Truven). In future work, we plan to incorporate lab data, which may further improve the model’s ability to predict overdose events. 
Third, due to the high cost of fine-tuning proprietary models like GPT-4o with a large number of input tokens, we limited our training dataset to 900 patients. In the future, we will incorporate additional training instances to investigate whether it could enhance the model’s predictive capabilities.
Finally, MedGemma fine-tuning was conducted only for the summarized visit formats, as the detailed visit formats produce input sequences that exceeded GPU memory capacity even with QLoRA. Future work should explore whether more memory-efficient fine-tuning strategies or hardware configurations can enable MedGemma fine-tuning on detailed formats. 

Another limitation of our model is that it relies on insurance claims data, which is not available in real time as soon as a visit concludes. The model predicts the risk of drug overdose within the next 7 or 30 days from an index date. However, details of the most recent visits leading up to that date may not yet be recorded in insurance claims, creating a gap in real-time prediction capabilities. Since our model primarily uses Merative data, which consists of consolidated claims that can take weeks to process, this delay prevents the model from utilizing the most up-to-date patient information. A potential solution for real-world deployment is to supplement Merative’s historical data with recent visit records from local hospitals or clinics (since the latest Merative record), before making predictions. By integrating these sources, the model will be able to generate more timely and reliable predictions. Thus, we emphasize  this limitation is not prohibitive to the implementation and utility of our methods.

\section*{Acknowledgement}
This work is supported by the U.S.~NIH  National Institute on Drug Abuse through grant R01DA057686.
The content is solely the responsibility of the authors and does not necessarily represent the official
views of the NIH.


\bibliography{custom}
\bibliographystyle{IEEEtran}

\section{Appendix}

\subsection{Number of Encounters}
\label{sec-encounter}
The number of prior encounters provided as input to the LLM plays a crucial role in determining its predictive performance, as it directly influences the amount of historical medical information available for decision-making. A larger number of encounters allows the model to capture more comprehensive details of the patient's medical history, potentially leading to more accurate predictions. However, increasing the number of encounters also results in longer input prompts, which can adversely impact the model’s ability to effectively process and utilize the information.

To investigate the effect of varying the number of encounters input to the LLM, we conducted a zero-shot experiment using different maximum encounter limits. For this experiment, we used the detailed descriptive prompt type and set the prediction window to 7 days. The results, presented in Table~\ref{table:zero-shot-different-visit}, show that the model's performance is lowest when only the most recent five encounters are used. As the number of encounters increased, the model's performance improved, indicating that more context enhances its ability to detect overdose risk patterns.
However, we observed a decline in performance when the maximum number of encounters was set to 40. While increasing the number of encounters provides more information for the LLM to make informed predictions, it also significantly lengthens the input prompt. For instance, in our experiment, the average token length for a prompt containing 20 encounters was 3070 tokens, whereas a prompt with 40 encounters reached an average of 6652 tokens. This excessive input length likely introduces challenges in the model's ability to focus on the most relevant information, leading to relatively lower performance. Our results suggest that when the input prompt becomes too long, the 
model may struggle to focus on the most critical information, reducing its effectiveness.

\begin{table}[h]
    \centering
    \begin{tabular}{ccccc}
        \toprule
        Pred. Window & Max \#visits & Precision & Recall & F1 \\
        \midrule
        \multirow{5}{*} {7 days}& 5 & 55.10 & 36.00 & 43.55 \\
        {} & 10 & 58.37 & 45.33 & 51.03 \\
        {} & 20 & 55.30 & 48.67 & 51.77 \\
        {} & 30 & 57.68 & 51.33 & 54.32 \\
        {} & 40 & 55.97 & 50.00 & 52.82 \\
        
        \bottomrule
    \end{tabular}
\caption{Zero-shot performance for different values of maximum number of visits considered.}
\label{table:zero-shot-different-visit}
\end{table}

From these experiments, we identified that using a maximum of 30 prior encounters strikes the optimal balance between providing sufficient historical context and maintaining a manageable input length. Consequently, we adopted 30 encounters as the default configuration for all subsequent experiments in this study.

\subsection{Descriptions Vs Original Medical Codes}

 LLMs are trained on diverse datasets, including biomedical and clinical text, and may therefore possess an inherent understanding of standardized medical coding systems such as ICD-9/ICD-10 for diagnoses, CPT/HCPCS for procedures, and other structured clinical terminologies. Unlike natural language descriptions, which provide detailed explanations of medical conditions and procedures, medical codes are compact representations that contain equivalent clinical information in a more structured format.

One key advantage of using medical codes over natural language descriptions is the significant reduction in input token length. Since medical codes consist of fewer tokens than their corresponding descriptions, replacing descriptions with codes substantially decreases the prompt size. For instance, when using the detailed descriptive prompt format with 30 prior encounters, the average input length is 5,510 tokens, whereas using only medical codes reduces the average token count to 4,147 — a reduction of approximately 25\%. This reduction not only optimizes computational efficiency and lowers API costs for GPT-4o, but is also particularly beneficial for MedGemma, where shorter prompts partially mitigate GPU memory requirements during fine-tuning. Both considerations are important for the practical deployment of LLMs in real-world clinical applications.

\begin{table*}[t]
    \centering
    \begin{tabular}{clcccc}
        \toprule
        \multirow{3}{*}{Model} & \multirow{3}{*}{Prompt Type} & \multicolumn{4}{c}{Model}\\
        \cmidrule{3-6}
        & & \multicolumn{2}{c}{GPT-4o} & \multicolumn{2}{c}{MedGemma-27B} \\
        \cmidrule{3-6}
        & & Exposed & Non-exposed & Exposed & Non-exposed\\
        \midrule
        \multirow{4}{*} {Zero-shot} & Detailed visit - descriptive & 73.67 & 88.67 & 10.67 &50.33\\
        {} & Detailed visit - med. code & 69.67 & 88.67 & 20.00 & 56.00\\
        {} & Summarized visit - descriptive & 69.00 & 96.67 & 44.67 &86.67 \\
        {} & Summarized visit - original code & 66.33 & 95.67 &46.67 &85.33 \\
        \midrule
        \multirow{4}{*} {Fine-tuned} & Detailed visit - descriptive & 88.00 & 90.00 & - & -\\
        {} & Detailed visit - med. code & 86.00 & 85.00 & - & -\\
        {} & Summarized visit - descriptive & 94.00 & 94.00 & 97.33 & 96.33\\
        {} & Summarized visit - original code & 95.67 & 95.00 & 93.67 & 97.33\\
        
        \bottomrule
    \end{tabular}
\caption{Accuracy of the models in predicting `No Overdose' for exposed vs. non-exposed instances within the control cohort.}
\label{table:exposed_control_acc}
\end{table*} 

To assess whether LLMs can effectively interpret medical codes and make predictions based on them, we evaluated both GPT-4o and MedGemma-27B-text-it using prompt formats that replace natural language descriptions with original medical codes. The results in Tables~\ref{table:LLM-zero-shot} and ~\ref{table:LLM-fine-tuned} indicate that performance using medical codes is comparable to that of descriptive inputs for both models. For example, for GPT-4o in the zero-shot 7-day prediction window, the F1-score drops only slightly from 54.32 (descriptive) to 52.69 (medical code), and a similarly small drop of approximately 3.5 points is observed for MedGemma under the same settings. This suggests that both models can effectively utilize structured clinical codes for overdose prediction, achieving performance levels similar to those obtained with descriptive inputs. This has implications for the practical deployment of LLMs in clinical settings, as using structured codes not only reduces computational costs but also ensures compatibility with electronic health record (EHR) systems, where codes are widely used for documentation and analysis. However, further research is needed to explore whether incorporating minimal supplementary context alongside medical codes could enhance interpretability and improve predictive accuracy.

\subsection{Sequential Encounter Information vs Aggregated Statistics}
Our experiments showed that traditional machine learning models, such as XGBoost and Random Forest, outperform LLMs in predicting overdose events when using detailed sequential encounter information. For LLM-based predictions, input data consists of textual descriptions of the prior n encounters presented in chronological order, where each encounter includes details on diagnoses, procedures, and medications, preserving the sequential nature of a patient's medical history. In contrast, traditional machine learning models process input data in a categorical format, where each unique diagnosis, procedure, or medication is treated as an independent feature. Instead of maintaining the chronological structure, these models aggregate information by counting the occurrences of each feature within the last n encounters, condensing a patient's history into a simpler numerical format that potentially enables better pattern recognition.

To assess if LLMs can leverage aggregated information similarly to traditional models, we transformed the structured feature vectors used by the baseline models into a textual format. Specifically, we represented the data as a sequence of key-value pairs, where the key corresponds to a medical feature (e.g., a diagnosis or procedure) and the value represents its frequency over the last $n$ visits. This approach allowed us to present the same aggregated statistical representation to the LLMs in a format they could process effectively. The results, presented in Table~\ref{table:LLM-zero-shot}, indicate that both LLMs underperform in the zero-shot setting when using the aggregated input format. However, in the fine-tuned setting, GPT-4o with aggregated statistics surpasses the baseline models (F1: 84.53 vs. 78.92 for XGBoost), demonstrating that LLMs can effectively learn to interpret aggregate data when fine-tuned accordingly. For MedGemma, fine-tuned results are available only for the summarized formats; the model achieves high precision and specificity (89.62 and 96.83, respectively, for the 7-day descriptive format) in this setting, suggesting that the aggregated format is well-suited to its post-fine-tuning behavior, even if recall remains lower than that of GPT-4o.

\subsection{Performance in the Exposed Group}
The exposed cohort is a subset of the control group where individuals have been exposed to opioids or stimulants but have not experienced a drug overdose. This cohort is especially challenging to predict, as individuals who are exposed to opioids or stimulants are generally at a higher risk of overdose. In our experiments, we evaluate the performance of both GPT-4o and MedGemma-27B-text-it   on this exposed cohort, comparing it to the non-exposed subset of the control group.
 Table~\ref{table:exposed_control_acc} shows the accuracy of both models in predicting `no overdose' for the exposed and non-exposed groups within the control cohort, using a 7-day prediction window. In the zero-shot setting, GPT-4o achieves lower accuracy on the exposed group (66.33--73.67\%) than on the non-exposed group (88.67--96.67\%) across all prompt formats, indicating that without task-specific training it tends to incorrectly associate opioid or stimulant exposure with a higher likelihood of overdose. MedGemma exhibits a far more pronounced version of this pattern, achieving only 10.67--46.67\% accuracy on the exposed group and 50.33--86.67\% on the non-exposed group in the zero-shot setting. This large gap suggests that MedGemma’s medical-domain pretraining causes it to treat opioid exposure as a strong signal of overdose risk, leading to substantial over-prediction on this subgroup without fine-tuning.

Fine-tuning substantially improves performance for both models on the exposed cohort. 
The best fine-tuned GPT-4o model achieves 95.67\% accuracy on the exposed group and 95.00\% on the non-exposed group, effectively eliminating the gap between the two subgroups. Fine-tuned MedGemma also shows strong improvement, achieving 97.33\% and 93.67\% accuracy on the exposed group for the summarized descriptive and medical code formats, respectively, and 96.33\% and 97.33\% on the non-exposed group. These results demonstrate that fine-tuning effectively corrects the tendency of both models to over-predict overdose risk for the exposed subgroup. Moreover, after fine-tuning, MedGemma performs comparably to, and in some configurations slightly better than, GPT-4o on this challenging cohort.

\subsection{Inference Cost}
 To estimate the cost of making a prediction with our trained model using the OpenAI API, we analyzed the cost per instance across different input prompt types for GPT-4o. Table \ref{table:api_cost} presents the average inference cost per instance for various prompt formats using 7-day and 30-day prediction windows, with up to 30 visits per patient. Prompts with detailed visit information and descriptive medical terms are the most expensive (\$0.0137 and \$0.0125 per instance for the 7-day and 30-day windows, respectively), as they contain a higher average number of tokens per patient. In contrast, replacing descriptive medical terms with their corresponding medical codes reduces the API cost by nearly 25\%. Furthermore, summarizing visit information lowers the cost even more (by nearly 73\%), as summary statistics require significantly fewer tokens to represent a patient’s medical history compared to detailed visit descriptions. MedGemma-27B-text-it, as an open-weight model deployed on local infrastructure, does not incur per-call API costs; however, it requires dedicated GPU resources for inference and fine-tuning, and the memory demands of its 27B-parameter architecture represent a different category of computational cost that should be considered for deployment.
 
\begin{table}[h]
    \centering
    \begin{tabular}{lcc}
        \toprule
        Prompt Type & 7 days & 30 days \\
        \midrule
        Detailed visit - descriptive & \$0.0137 & \$0.0125 \\
        Detailed visit - med. code & \$0.0102 & \$0.0101 \\
        Summarized visit  - descriptive & \$0.0037 & \$0.0034 \\
        Summarized visit - original code & \$0.0031 & \$0.0029 \\
        
        \bottomrule
    \end{tabular}
\caption{The average cost in USD per instance for making predictions across different prompt types and prediction windows.}
\label{table:api_cost}
\end{table}

\end{document}